\documentclass{article}

\usepackage{PRIMEarxiv}

\usepackage[utf8]{inputenc} 
\usepackage[T1]{fontenc}    
\usepackage{hyperref}       
\usepackage{url}            
\usepackage{booktabs}       
\usepackage{amsfonts}       
\usepackage{nicefrac}       
\usepackage{microtype}      
\usepackage{lipsum}
\usepackage{fancyhdr}  
\usepackage{algorithm,algpseudocode}
\usepackage{caption}
\usepackage{graphicx}       
\graphicspath{{media/}}     

\title{Optimizing Active Learning for Low Annotation Budgets}

\author{Umang Aggarwal$^{1,2}$, Adrian Popescu$^{1}$, Céline Hudelot$^{2}$\\
\small{(1) Université Paris-Saclay, CEA, Département Intelligence Ambiante et Systèmes Interactifs}\\
\small{(2) 
Université Paris-Saclay, CentraleSupélec, Mathématiques et Informatique pour la Complexité et les Systèmes} \\
\small{91191 Gif-sur-Yvette, France}\\
{\tt\small {umang.aggarwal,adrian.popescu}@cea.fr,celine.hudelot@centralesupelec.fr}
}

\begin{document}
\maketitle

\begin{abstract}
When we can not assume a large amount of annotated data , active learning is a good strategy. It consists in learning a model on a small amount of annotated data (annotation budget) and in choosing the best set of points to annotate in order to improve the previous model and gain in generalization.
In deep learning, active learning is usually implemented as an iterative process in which successive deep models are updated via fine tuning, but it still poses some issues. 
First, the initial batch of annotated images has to be sufficiently large to train a deep model.
Such an assumption is strong, especially when the total annotation budget is reduced.
We tackle this issue by using an approach inspired by transfer learning.
A pre-trained model is used as a feature extractor and only shallow classifiers are learned during the active iterations.
The second issue is the effectiveness of probability or feature estimates of early models for AL task. Samples are generally selected for annotation using acquisition functions based only on the last learned model.  
We introduce a novel acquisition function which exploits the iterative nature of AL process to select samples in a more robust fashion. 
Samples for which there is a maximum shift towards uncertainty between the last two learned models predictions are favored.
A diversification step is added to select samples from different regions of the classification space and thus introduces a representativeness component in our approach.
Evaluation is done against competitive methods with three balanced and imbalanced datasets and outperforms them.
\end{abstract}


\section{Introduction}

Deep neural networks learn complex feature representations by exploiting large datasets to optimize a large number of parameters. 
Deep supervised learning strategies have given impressive results in multiple domains, but their effectiveness is governed by the size of the annotated dataset. 
Large annotated datasets are not readily available in many applications.
Active Learning (AL)~\cite{Settles10activelearning} reduces the cost of creating a labeled dataset by selecting, through an acquisition function (AF) the most relevant samples to train the model.
The effectiveness of the active learning process depends on the strategy deployed for sample selection.

A large number of sample acquisition functions have been proposed to improve the AL process.
One mainstream strategy is to maximize the informativeness by selecting the samples on which the model gives most uncertain probability estimates~\cite{DBLP:conf/cvpr/BeluchGNK18,DBLP:conf/aaai/CulottaM05,DBLP:conf/icdm/SchefferDW01,Shannon1948}. 
Alternatively, sample acquisition can be designed to select diverse samples which maximize the likelihood between distribution of labeled and unlabeled samples~\cite{DBLP:conf/icml/DasguptaH08, DBLP:conf/icmla/LiGC12,DBLP:conf/iclr/SenerS18}. 
Informativeness and diversity are not easy to optimize jointly but, since they convey complementary cues, their combination is tackled in works such as~\cite{ash2020deep,DBLP:journals/pami/ChakrabortyBSPY15} .

Active learning is generally implemented in an iterative fashion, with a new batch of samples being annotated in each iteration~\cite{DBLP:conf/cvpr/BeluchGNK18,DBLP:conf/icml/GalIG17,Settles10activelearning}. 
Models are retrained at the end of each iteration in order to incorporate the newly annotated samples which improve the models' predictive power.
One open problem in AL is that a relatively large amount of samples needs to be annotated before a viable initial deep model can be learned efficiently~\cite{gao2020consistency,konyushkova2017learning,Settles10activelearning,DBLP:conf/cvpr/ZhouSZGGL17}.
The size of the initial labeled dataset should be such that a model trained on the dataset provides reliable uncertainty measures and feature representations for the AL acquisition functions to outperform random sampling. 
This issue is further exacerbated for deep learning models that are quite data intensive. 
The size of the initial labeled dataset depends on the deep architecture used, the complexity of the problem and the acquisition function.
In our work, we explore training of a Support Vector Machine (SVM) classifier over fixed representation as an alternate to the dominant fine tuning scheme used in recent works~\cite{DBLP:conf/cvpr/BeluchGNK18,DBLP:conf/icml/GalIG17,gao2020consistency}.
This type of approach, which instantiates transfer learning, has been shown to be effective for active learning over imbalanced datasets~\cite{Aggarwal_2020_WACV}.
\begin{figure*}[]

\includegraphics[width=0.99\textwidth]{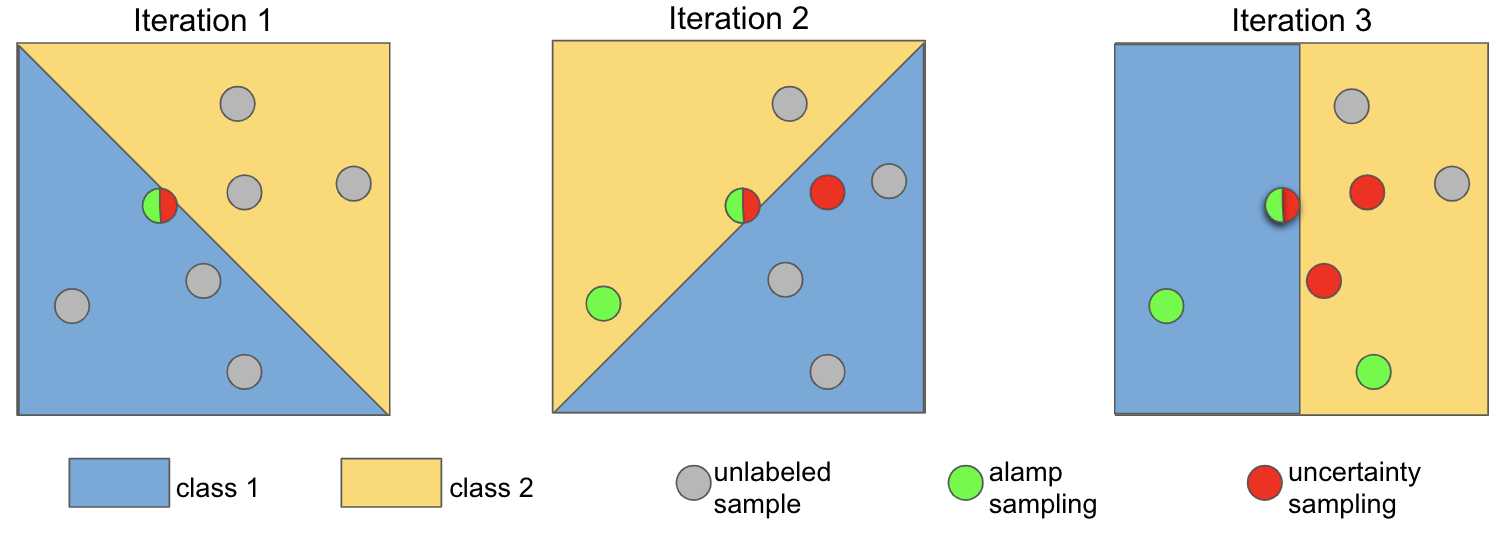}

\vspace{-0.5em}
    
    \caption{Illustration of $alamp$, the proposed method, for a two class problem. Blue and yellow regions are the class boundaries learnt by classifier. One sample is selected at each iteration.
    Uncertainty sampling selects the sample (red) which is closest to the class boundary.
    At Iteration 1, $alamp$ selects the same sample as  sampling, since it has is no access to previous model. At Iteration 2,
    $alamp$ selects the sample which gives most certain prediction (i.e. is furthest away from decision boundary) at Iteration 1 and gives the most uncertain prediction (i.e. is closest to decision boundary) at Iteration 2. Similarly at Iteration 3, $alamp$ selects the sample with maximum shift from certainty to uncertainty }
    \label{fig:method}
\vspace{-3mm}
\end{figure*}

Our main contribution is a new measure of informativeness which integrates predicted probabilities in successive AL iterations. 
Samples whose prediction states move from certain to uncertain between two iterations are prioritised. The underlying intuition here is that such samples encode information which is missing from the models and should be integrated into them. The measure is illustrated for a two-class problem in Figure ~\ref{fig:method}.
This view of informativeness is broader than the one incorporated in current uncertainty-based sampling which only considers the decision boundaries of the current iteration.

However, the proposed sample AF shares a limitation with existing informativeness-based functions in that it could suffer from a lack of representativeness~\cite{Settles10activelearning}. 
The second contribution is to add a representativeness dimension in the proposed acquisition function.
Representativeness is modeled via the introduction of a diversification procedure.
The proposed informativeness measure prioritizes samples with high certainty at the previous iteration. 
The diversification procedure exploits this fact to select samples with different class predictions in the previous iteration. 
This leads to selection of samples from different uncertain regions distributed across the classification space provided by the model classifier. 

The two contributions are well-suited for use with a shallow classifier trained over fixed representations.
The constant nature of representation helps the proposed measure to effectively evaluate the distance of the samples to the classifier boundary in the preceding and the current iterations.
We show that training a shallow classifier over fixed representations gives higher performance than fine-tuning CNN models, especially for smaller budgets. 
This is important insofar fixed representations can be exploited from the very beginning for the AL task where the classical fine tuning estimates are shown to be unstable~\cite{gao2020consistency}.
The experiments are performed with three balanced datasets: Cifar100, Food-101 and a subset of ImageNet classes which are not part of ILSVRC. We also test our method for imbalanced datasets by creating imbalanced version of Cifar100 and Food-101 and using MIT-Indoor67 which is naturally imbalanced.
ILSVRC itself is used to create the fixed representation. 
The proposed method is compared to a series of strong baselines. 
The results indicate that it outperforms the baselines in most of the evaluated configurations.

\section{Related works}
Active learning was studied thoroughly in the pre-deep learning period~\cite{Settles10activelearning} and also recently in deep learning~\cite{ash2020deep, DBLP:conf/cvpr/BeluchGNK18,DBLP:conf/icml/GalIG17, DBLP:conf/iclr/SenerS18}. 
Several heuristics have been designed to select the most important samples that help in improving the model. They fall in two main categories. One set of methods focuses on informative samples that would provide new information to the model. 
Uncertainty-based acquisition functions select samples for which the model gives the least sure predictions. 
Some of the most representative AFs proposed in literature include: entropy~\cite{Shannon1948}, least confidence~\cite{DBLP:conf/aaai/CulottaM05} and margin sampling~\cite{DBLP:conf/icdm/SchefferDW01}.
Margin Sampling prioritizes samples for which the model is least certain between the top 2 class predictions. 
It is shown to be effective in AL task as it relates well to selecting samples based on distance from the decision boundary for shallow clasifier such as SVMs~\cite{tong2001support,brinker2003incorporating}.
It has also been used in imbalanced learning context with SVM classifier to select samples which are closest to the decision boundary of the classifier~\cite{attenberg2013class, ertekin2007active, zhu2007active}. 
Hence, given the use of SVM training scheme in our work we study margin sampling as the main uncertainty measure.
The main limitation of informativeness-based approaches is that they might fail to capture the overall sample distribution because they provide no guarantee of a thorough exploration of the representation space.
As a result, they are susceptible to provide a biased selection. 

Another set of techniques are AFs on the objective of representativeness. These approaches select a diverse set of samples for annotation which are most likely to represent the unknown distribution. Several approaches based on clustering methods such as K-means or hierarchical clustering have been introduced in active learning~\cite{DBLP:conf/icml/DasguptaH08}. These methods select a diverse set of sample by sampling from different clusters.
Coreset~\cite{DBLP:conf/iclr/SenerS18} is a recent method that solves the K-center problem by selecting the samples with maximum distance to its closest labeled neighbour. 

The joint use of informativeness and representativeness can be challenging due to different nature of the two selection strategies. 
However, they convey complementary information and set of approaches tackled their combination in order to select samples which are both representative and uncertain~\cite{han2016local,hsu2015active, wei2015submodularity}. 
In~\cite{hsu2015active}, a strategy is proposed to select using either uncertainty sampling or representative sampling by a sequential decision process.
Badge~\cite{ash2020deep} is a recent work which exploits the gradient of the embedding of the last layer of the model. A K-means++ clustering is implemented on the gradient embedding to select diverse samples.
It is argued that kmeans++ selects samples with a large gradient value and thus the samples are also informative. We also try to combine the two objectives by selecting uncertain samples from diverse regions of multi-class classifier but using a different strategy.

A direction of research tries to improve the quality of the uncertainty estimates. Bayesian probabilities were introduced as better estimate of uncertainty by combining probabilities of several runs of model~\cite{DBLP:conf/icml/GalIG17}. The dropout parameter is used during inference to ensure variation in model probabilities. Ensemble models have also been used to acquire multiple probability estimates. In~\cite{DBLP:conf/cvpr/BeluchGNK18}, an ensemble of model snapshots is created by using a cyclic learning rate.
This design choice is important in order to limit the computational effort need to create the ensemble. 
In our work, we derive multiple probabilities by exploiting the iterative nature of AL cycle without additional computational cost.

Several works add the semi-supervised learning objective to active learning by using the unlabelled data to train the model. 
An approach to find the optimal acquisition function using feature density matching between unlabelled dataset and weakly supervised validation data is presented in~\cite{Gudovskiy_2020_CVPR}. 
Inspired by MixMatch~\cite{berthelot2019mixmatch}, consistency based semi-supervised learning~\cite{gao2020consistency} has been used for AL task by selecting samples which  give inconsistent predictions for different permutations. 
Our work has similar approach of prioritising samples with inconsistent predictions, but at different iterative steps.  
Semi-supervised learning is difficult to generalize, particularly for very low budgets which is in focus here. 
More importantly, most of the methods in this category use an additional labeled validation set to optimize the parameters.
It is not realistic to assume that such a set exists at the beginning of the AL process. 
Moreover, these works come with an added complexity in terms of computational resources.
Hence they are not directly comparable with shallow classifier used here.

An approach which exploits the model learned at previous acquisition step via knowledge distillation is presented in~\cite{yunweight}. They use the previous model to fine tune the weight decay parameter and provide effective regularization. The contribution is complementary to ours. The key difference is that we use the probability distribution from the previous iteration to propose a measure of informativeness.

An initial annotated dataset is necessary to kick-start the iterative AL cycle. The most common approach is to annotate a small randomly selected subset~\cite{DBLP:conf/cvpr/BeluchGNK18,DBLP:conf/icml/GalIG17,DBLP:conf/iclr/SenerS18}. 
The size of the initial set is critical in AL.
This is particularly the case for large deep neural networks which might not provide reliable and stable probability estimate at the start of AL cycle. 
An approach to mitigate this problem for DNN was presented in~\cite{Coleman2020Selection}. 
They showed that a smaller proxy network can be used for sample selection as at smaller budget, a smaller model provides better estimates compared to a larger model. The large model is trained when sufficient samples are selected.
A transfer learning perspective was introduced in AL to select the initial batch of samples~\cite{Aggarwal_2020_WACV}. The probability estimates for the unlabelled dataset is derived from a pre-trained model to select a diverse and balanced set of samples. 
Similar to~\cite{Aggarwal_2020_WACV}, we test the use of shallow classifier on top of pre-trained features as compared to the classical fine-tuning approach.

\section{Problem Description and Baselines}

We consider an unlabelled dataset $\mathbb{D}^{U}$ with samples $x_i \in \mathcal{X}$ for $i = [1..u] $, i.i.d realizations of random variables $\mathcal{X}$ drawn from the distribution $\mathbb{P}$ where $\mathcal{X}$ is the instance space. 
In an iterative AL setting, a total budget $b$, with $b < u$, is allocated for manual labeling in $t$ iterations.
The process starts by randomly selecting a small subset of $\mathbb{D}^{U}$ for annotation to create the initial labeled dataset $\mathbb{D}_{0}^{L}$ with $x_j,y_j \in \mathcal{X} \times \mathcal{Y}$  for $j = [1..\frac{b}{t}]$.

Here 
$\mathcal{Y} = \{y_1,...,y_n\}$ is the set of $n$ class labels. 
$\mathbb{D}_{0}^{L}$ is used to train an initial model $\mathcal{M}_{0}$ which provides the probability estimates $\mathcal{P}_{0}$  or the feature embeddings $\mathcal{F}_{0}$ required by an acquisition function AF to estimate the importance of samples for the next iteration.
Afterwards, at iteration step $k$, for $k = [1..t-1]$, a batch of samples of size $\frac{b}{t}$ is selected for labeling from $\mathbb{D}_{k}^{U} = \mathbb{D}^{U} \setminus \mathbb{D}_{k-1}^{L}$, and added to $\mathbb{D}_{k-1}^{L}$ to update the labeled subset $\mathbb{D}_{k}^{L}$. 
$\mathbb{D}_{k}^{L}$, the labeled set at iteration $k$, is then used to learn the model $\mathcal{M}_{k}$.

We test two training methods which are based either on deep model update or on a pretrained model followed by a shallow learning of classifiers. 
Both schemes exploit a pretrained model $\mathcal{M}_{S}$ learned over a generic dataset.
The first method exploits fine tuning, which is the usual way in which deep models are updated in iterative AL~\cite{DBLP:conf/cvpr/BeluchGNK18,DBLP:conf/icml/GalIG17,DBLP:conf/iclr/SenerS18}.
Deep models are fine tuned for each AL iteration using $\mathcal{M}_{S}$ and $\mathbb{D}_{k}^{L}$.
The second method assumes that, although $\mathcal{M}_{S}$ is trained on a separate dataset, the knowledge it encapsulates can be transferred to $\mathbb{D}^{U}$.
Feature embeddings $\mathcal{F}_{S}$ from $\mathcal{M}_{S}$ are used to train shallow classifiers during AL iterations.

\subsection{Baselines}
We compare our method to four competitive baselines. The first three are classical methods representative for random, informativeness-based, and diversity-based selection respectively.
The fourth is $badge$, a recently introduced method which combines the informativeness and representativeness objectives ~\cite{ash2020deep}. This baseline is suited only in the fine tuning training scheme.

\subsubsection{Random sampling}
The first baseline $rand$ consists in a random selection of samples for annotation. The initial subset to start the AL iterative cycle is selected using random sampling.
While very simple, random sampling is a strong baseline in active learning~\cite{DBLP:conf/cvpr/BeluchGNK18,DBLP:conf/iclr/SenerS18,Settles10activelearning}.

\subsubsection{Margin sampling}
Uncertainty based methods aim to maximize informativeness and thus focus on samples that the model classifies with the most difficulty.
They are deployed using the outputs of the classification layer~\cite{Settles10activelearning}.
Margin based AL is shown to be an effective uncertainty measure for SVM classifier ~\cite{tong2001support,ertekin2007active,attenberg2013class}.
It is defined as:
\begin{equation}
 marg(x) =   (P(y = c^1|x) - P(y = c^2|x))   
 \label{eq:marg}
\end{equation}
where $c^1,c^2$ are the top-2 predicted classes for test sample $x$ at iteration $k$.

We consider $\textrm{Marg}_{\textrm{sort}} (\mathbb{D}_{k}^{U})$ a permutation of the set $\mathbb{D}_{k}^{U}$ by ordering its element by increasing value of $marg$ . $\mathbb{D}_{k+1}^{L}$ is obtained by the union of $\mathbb{D}_{k}^{L}$ with the first $\frac{b}{t}$ samples of $\textrm{Marg}_{\textrm{sort}} (\mathbb{D}_{k}^{U})$.
The baseline obtained is noted $marg$.


\subsubsection{Coreset}
Diversity based methods select a subset of $\mathbb{D}^U$ so as to ensure an optimal coverage of the representation space.
Coreset~\cite{DBLP:conf/iclr/SenerS18} takes a minmax view to select the diverse samples.
It is defined as:
\begin{equation}
  core(\mathbb{D}_{k}^{U},\mathbb{D}_{k}^{L}) = 
    \max_{\forall x_u \in \mathbb{D}_{k}^{U}} \min_{x_l \in \mathbb{D}_{k}^{L} } d(F(x_u),F(x_l))  
  \label{eq:core}
\end{equation}
where $core(\mathbb{D}_{k}^{U},\mathbb{D}_{k}^{L})$  returns a sample from unlabeled dataset $\mathbb{D}_{k}^{U}$ using the labeled dataset  $\mathbb{D}_{k}^{L}$,  $d(F(x_u),F(x_l))$ is the distance between a labeled point $x_l$ from $\mathbb{D}_{k}^{L}$ and an unlabeled point $x_u$ from $\mathbb{D}_{k}^{U}$. $F$ is the feature extractor. For the SVM training, a classifier is learned with $F$ = $F_S$ given by the pretrained model $\mathcal{M}_S$. 
For the CNN fine tuning, the feature extractor is given by $\mathcal{M}_{k-1}$, the model available from the latest iteration. 
In Equation~\ref{eq:core}, the unlabeled sample which is furthest away from its closest labeled sample is selected.  
The baseline AF obtained is noted $core$.

\section{Method}
Classical AFs select the samples based on either the probability or feature estimates from a single model. We hypothesize that using the model predictions from the model learned in the previous active learning cycle can lead to more efficient sampling. 
The informativeness of samples is ascertained by taking into account the change in their probability distribution in successive iterations. 
The strategy prioritizes samples which were predicted with high certainty in the previous iteration but which gives uncertain prediction in the current model.
We derive an analogy to a student who gave confident response to a question, but becomes uncertain after learning some more information. Knowing the true answer should benefit the student and provide relevant missing information.  
The strategy is well-suited for the low budget setting where the batch size is generally small. The update of model with large number of samples could make the precedent model less relevant. 
Focusing on the unlabeled samples on which the model becomes uncertain in its predictions adds a novel component of uncertainty in the selection process.
In cases where the sample was correctly predicted, selecting these samples allows the model to focus on samples that it is most likely to forget.
Alternatively, even if a sample was predicted incorrectly, the measure allows to select more difficult samples which can improve the generalization ability of the model.
Hence, knowing the labels of these samples should be informative.

\subsection{alamp: active learning with asynchronous model predictions}

We present here the formulation of $alamp$ which allows to select the samples with the maximum shift from certainty to uncertainty between the two iterations.
Note that any of the uncertainty measures can be used in $alamp$.
The definition of the method which exploits margin sampling as basic acquisition function is:
\begin{equation}
 alamp(x) = \frac{marg_{k-1}(x) - marg_{k}(x)}{marg_{k-1}(x) + marg_{k}(x)}
 \label{eq:iter_ms}
\end{equation}
with $marg_k : \mathbb{D}_{k}^{U} \to \mathbb{R}$, the margin function as in equation 1 defined on the set $\mathbb{D}_{k}^{U} $.

The score takes a normalized min-max view of uncertainty allowing to select samples with maximum change (certainty to uncertainty) between the iteration. 
The sample with higher score is selected. The numerator gives the difference in the certainty between the previous and current iteration. The denominator normalizes the certainties to ensure that for samples which same absolute difference in certainty(numerator) one with lower sum of certainties is selected. This allows the score to select samples with maximum relative shift towards uncertainty.

We consider $\textrm{Alamp}_{\textrm{sort}} (\mathbb{D}_{k}^{U})$ a permutation of the set $\mathbb{D}_{k}^{U}$ by ordering its element by decreasing value of $alamp$ . $\mathbb{D}_{k+1}^{L}$ is obtained by the union of $\mathbb{D}_{k}^{L}$ with the first $\frac{b}{t}$ samples of $\textrm{Alamp}_{\textrm{sort}} (\mathbb{D}_{k}^{U})$.

The method has low supplementary memory requirements since it stores the probability distributions at each iterative step.
At the first iterative step, we have access to only the probability estimate $\mathcal{P}_{0}$ from the initial model $\mathcal{M}_{0}$, which is trained over the initial randomly selected dataset $\mathbb{D}_{0}^{L}$, thus the selection is based on the uncertainty criteria in the first iterative step. 

\subsection{alamp-div}

$alamp$ inherits the limitations of the uncertainty-based method used in its definition in terms of sample representativity. 
We introduce a variant of the method, named $alamp-div$ which selects informative samples from different regions of the classification space.
$alamp-div$ is described in Algorithm~\ref{alg:div}.
$alamp$ sorts the samples based on the difference of certainty measure of the sample on the previous iteration and current iteration. The samples can therefore be assigned to pseudo classes according to the class predicted in the previous iteration. 
The selection process is driven toward selecting the same number of samples from each pseudo class so to aim for representativeness and balance across classes. 
This enables the selection of informative samples across a diverse set of classifier boundaries in a multi class problem.  

\begin{algorithm}
\caption{Diversification algorithm}
\label{alg:div}
\begin{algorithmic}[1]
\State $U$: a list of unlabeled samples sorted according to AF
\State $k$: current iterative step 
\State $top$: a dictionary which assigns top class prediction in the iteration $k-1$ to all samples in $U$ 
\State $b$ : budget of samples to be selected
    \Procedure{div}{$U$, $top$, $b$}

    \State Build $L$: list of samples selected from $U$ of length $b$
    \While {len($L$)$\leq$ $b$} 
    \State seenclasses = empty list : reinitialize memory of classes 
    \For{each item $i$ in $U$}
    \State $topclass =$ top[i]    :predicted class at iteration $k-1$ for sample $U[i]$
        \If{$topclass$ not in $seenclasses$ } 
            \If{$i$ not in $L$}
                \State add sample $i$ in $L$
                \State add $topclass$ in $seenclasses$
            \EndIf    
        \EndIf    
    \EndFor
    \EndWhile
    \State $L = L[0:b]$ 
    \State \textbf{return} $L$
    \EndProcedure
\end{algorithmic}
\end{algorithm}

\section{Experiments}
We first describe the experimental setup for the transfer and fine-tuning training schemes that are tested.
Then, we describe the evaluation datasets. 
Finally, we present the results and their analysis.

\subsection{Setup}
The experimental setup is designed to focus on the small annotation budgets, which is most challenging for AL. 
In our experiment, we use 200 samples for the transfer and fine-tuning initial AL budget. 
The AL process is then run for 15 iterations with 200 samples selected at each iteration. 
The total budget at the end of the process includes 3200 samples.
Additional experiments with a higher budget setting are provided in the supplementary material.

We experiment with two training schemes. For comparability, a ResNet-18 architecture~\cite{DBLP:conf/cvpr/HeZRS16} is used as backbone of both of the training schemes tested here.
The first scheme, noted $FT$, is based on fine-tuning and mirrors the dominant approach in existing deep AL works~\cite{DBLP:conf/icml/GalIG17,DBLP:conf/iclr/SenerS18}.
We fine-tune the pre-trained model for 80 epochs. All the parameters are optimised using stochastic gradient descent with Nesterov momentum of 0.9. The initial learning rate is 0.01 and is reduced by a factor of 10 when the train error rate plateaus for 10 epochs. We use a weight decay parameter of 0.001. 
The models are trained using the Pytorch framework. Thresholding~\cite{DBLP:journals/nn/BudaMM18} which is shown to be effective to mitigate imbalance for deep models is used in experiments with imbalanced datasets.
The experiments are repeated for 5 runs and the average performance is reported.

$FT$ can be suboptimal when the budget is small enough to optimize the large number of parameters of the DNN. 
An alternate to $FT$ is to learn a SVM classifier over the features of the pre-trained model. Transfer learning scheme, noted as $\textrm{SVM}$ here, has been proven to be beneficial when the number of annotated AL samples is limited~\cite{Aggarwal_2020_WACV}. In our work, we exploit the features of a pre-trained model on $ILSVRC$~\cite{DBLP:journals/ijcv/RussakovskyDSKS15} dataset. 
The scikit-learn implementation of SVC classifier is used with standard default parameters. The cost sensitive SVM implementation is used for imbalanced datasets.
The regularization parameter is selected using a cross-validation on the training data.  
While sub-optimal, the use of training data for validation necessary because of data scarcity specific to AL.
The $SVM$ training scheme is deterministic once the initial subset is selected.

\subsection{Datasets}
The acquisition function and the training schemes are tested on three publicly-available image classification datasets:
\begin{itemize}
    \item $Cifar100$~\cite{Krizhevsky09learningmultiple} is designed for coarse-grained object classification. 
    \item $Food-101$~\cite{bossard14} is focused on fine-grained food recognition
    \item $IMN-100$  is a subset of ImageNet which includes fine-grained classes (i.e. ImageNet leaves). Note that the intersection between $IMN-100$ and $ILSVRC$ is empty.
\end{itemize} 
The three datasets are balanced and their main statistics are provided in Table ~\ref{tab:dataset}.
In addition, we test our methods on three imbalanced datasets. 
$MIT-67$ ~\cite{DBLP:conf/cvpr/QuattoniT09} is designed for indoor scene recognition and is imbalanced.
We also induce imbalance in $Cifar100$ and $Food-101$. 
An imbalance induction procedure was applied to $Cifar100$ and $Food-101$ for all datasets to have similar imbalance ratio for standardized comparison.
The imbalance ratio is defined as $ir = \frac{\sigma}{\mu}$, with $\sigma$ standard deviation and $\mu$ the mean of images per class in the dataset.
The main statistics of the obtained datasets are provided in Table~\ref{tab:dataset_imbal}. 

\begin{table}[]
    \begin{center}
    \resizebox{0.45\textwidth}{!}
    {
    \begin{tabular}{|c||c|c|c|}
        \hline
         Dataset & Class & Train images/class & Test images/class \\ \hline \hline
         Cifar100 &  100  & 500  & 100  \\ \hline 
         Food-101 &  101  & 750  & 250  \\ \hline
         IMN-100   &   100  & 1000  & 200  \\ \hline
    \end{tabular}
    }
    \end{center}
    \caption{Dataset statistics. }
    \label{tab:dataset}
\vspace{-3mm}
\end{table}

\begin{table}[]
    \begin{center}
    \resizebox{0.5\textwidth}{!}
    {
    \begin{tabular}{|c|c|c|c|c|c|}
        \hline
         Dataset & Class & Train Images & Mean($\mu$) & Std($\sigma$) & $ir$  \\ \hline
         Food-101 &  101  & 22956  & 227.28 & 180.31   & 0.793  \\ \hline
         CIFAR-100 &  100  & 17168  & 171.68 & 126.98   & 0.740  \\ \hline 
         MIT-67   &   67  & 14281  & 213.15 & 168.16   & 0.789  \\ \hline
    \end{tabular}
    }
    \end{center}
    \caption{Dataset statistics. $ir$ is the imbalance ratio.}
    \label{tab:dataset_imbal}
\vspace{-3mm}    
\end{table}

\begin{figure*}[ht]

\includegraphics[width=0.33\textwidth]{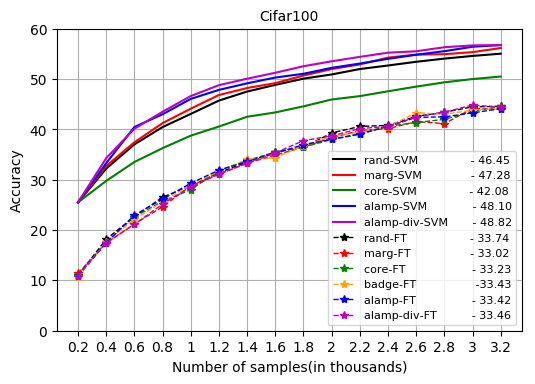}
\includegraphics[width=0.33\textwidth]{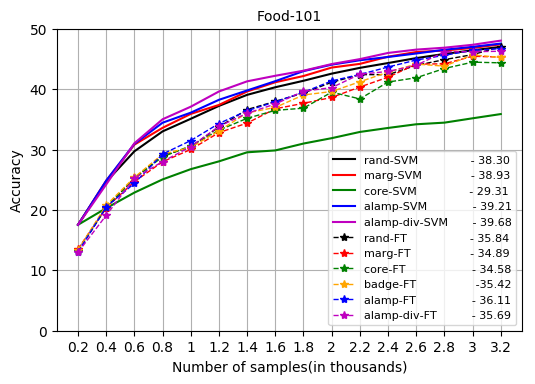}
\includegraphics[width=0.33\textwidth]{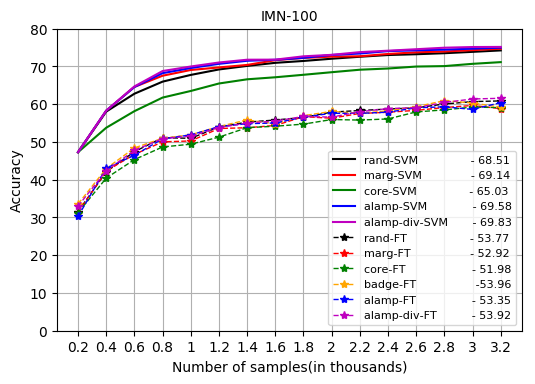}

\vspace{-1em}
    \caption{Iterative active learning performance with $SVM$ and $FT$ training schemes at each of 15 iterative steps for \textbf{balanced} datasets Cifar100, Food-101 and IMN-100 with initial budget of 200 and total budget of 3200. 200 samples added at each iteration  \textit{Best viewed in color}.}
    \label{fig:acc}
\vspace{-3mm}
\end{figure*}

\begin{figure*}[ht]

\includegraphics[width=0.33\textwidth]{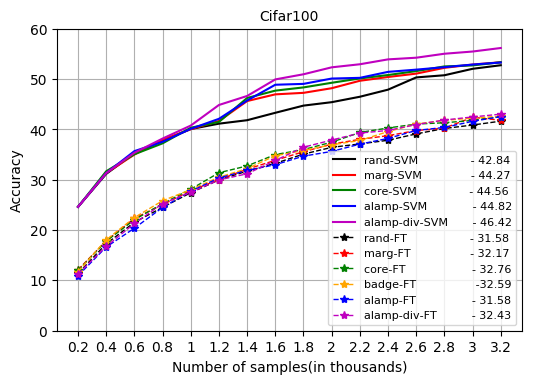}
\includegraphics[width=0.33\textwidth]{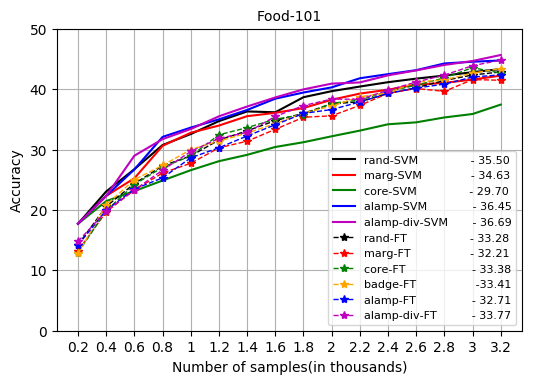}
\includegraphics[width=0.33\textwidth]{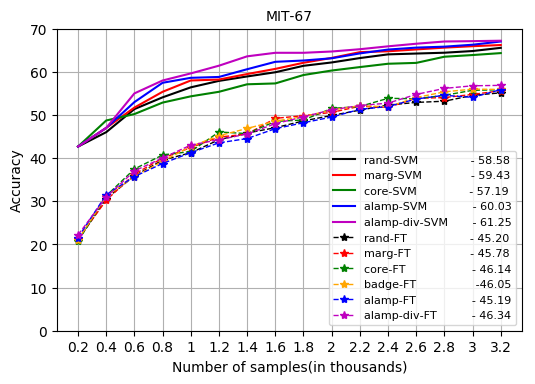}

\vspace{-1em}
    \caption{Iterative active learning performance accuracy with $SVM$ and $FT$ training schemes at each of 15 iterative steps for \textbf{imbalanced} datasets Cifar100, Food-101 and MIT-67 with initial budget of 200 and total budget of 3200. 200 samples added at each iteration  \textit{Best viewed in color}.}
    \label{fig:acc_im}
\vspace{-3mm}
\end{figure*}

\begin{figure*}[ht]

\includegraphics[width=0.33\textwidth]{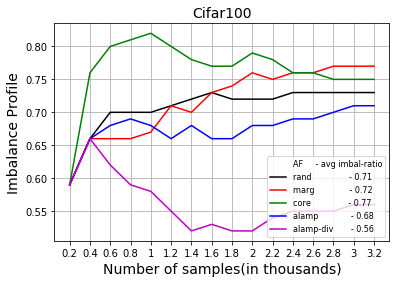}
\includegraphics[width=0.33\textwidth]{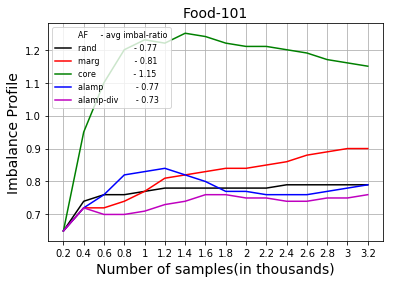}
\includegraphics[width=0.33\textwidth]{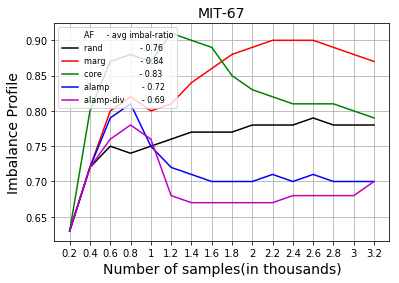}

\vspace{-1em}
    \caption{Imbalance profile (bottom) with $SVM$ training schemes at each of 15 iterative steps for \textbf{imbalanced} datasets Cifar100, Food-101 and MIT-67 with initial budget of 200 and total budget of 3200. 200 samples added at each iteration  \textit{Best viewed in color}.}
    \label{fig:ip_im}
\vspace{-3mm}
\end{figure*}

\begin{figure}[]
\centering
\includegraphics[width=0.48\textwidth]{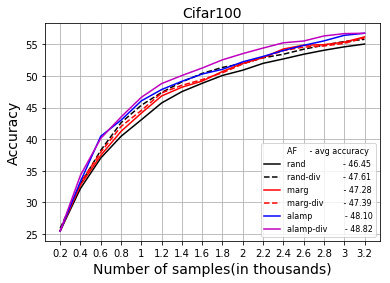}
\vspace{-1em}
    \caption{Iterative active learning performance for diversification applied to $rand$ and $marg$ with  $SVM$ training schemes at each of 15 iterative steps for Cifar100 with initial budget of 200 and total budget of 3200. 200 samples added at each iteration.}
\label{fig:div}
\vspace{-3mm}
\end{figure}

\subsection{Analysis of results}

The results obtained for the three balanced datasets using the baselines and the proposed methods are presented in Figure~\ref{fig:acc}.
Note that we provide both the detailed evolution of accuracy across AL iterations and the averaged performance of each method. 
Globally, $alamp$ and $alamp-div$ provide the best performance across the three datasets for $SVM$ training scheme. 
$random$ acts as a strong baseline, especially for $FT$ training scheme, where none of the methods that are tested can outperform $random$. Further the performance of $SVM$ scheme is clearly higher than that of the usual $FT$ scheme.
This is an interesting result which is analyzed in detail in Subsection~\ref{subsec:cold}.

Here we discuss the performance of different AFs in the $SVM$ training scheme. 
The average accuracy gain for the entire AL cycle is 2.4 , 1.3 and 1.3 points for $alamp$ compared $rand$ for balanced versions of $Cifar100$, $Food-101$ and $IMN-100$ respectively.
More interesting from a practical perspective, the number of samples required for achieving 50 percentage of accuracy for $Cifar100$ is 1800 with $random$ or $marg$, 1600 for $alamp$ and 1400 for $alamp-div$.
Similarly for $Food-101$, 40 percent accuracy is reached with around 1600 samples for $random$, 1400 samples with $alamp$ and $marg$ and 1200 samples for $alamp-div$. $alamp-div$ avoids the annotation of 400 extra samples as compared to $random$ to achieve 50 percent accuracy for $Cifar100$ and 40 percent accuracy for $Food-101$. 
The performance gain is more limited for $IMN-100$. Overall accuracy for this dataset is already quite high with $SVM$ training scheme. 
This is an expected result, since $IMN-100$ is closest to the $ILSVRC$ dataset used to train the source model. $alamp-div$ and $alamp$ are still the best methods with 70 percent accuracy attained with 1000 samples with $alamp$ and $alamp-div$, while $rand$ and $marg$ require 1200 samples.  

$marg$ outperforms $random$ in the $SVM$ training scheme, showing that $SVM$ classifier provides reliable uncertainty estimates even at low budgets.
In our experiments, $marg$ becomes competitive to $alamp$ at large budgets when the uncertainty estimates become stronger. 
This is explained by the fact that, as the accuracy of the model increases, uncertainty measures becomes more important to find the missing information.
This is the case of $IMN-100$, which has the highest overall accuracy among the three dataset tested. 
The accuracy of $IMN-100$ is around 50 percent at the start of AL cycle and $marg$ is more competitive for $IMN-100$ than for the other two datasets.
 
$core$ has suboptimal performance in the $SVM$ training scheme for all datasets. This is particularly the case of $Food-101$, which is most different from the $ILSVRC$ dataset used as feature extractor. It gives comparable performance to other AFs in the $FT$ training scheme where the feature extractor is updated along with the classifier. 
$badge$ is suited only for $FT$ training scheme as it requires the gradients on the features. In $SVM$ training scheme, the features are fixed and hence it is not possible to test $badge$. In the $FT$ training scheme, $badge$ also  fails to provide any significant improvement over $random$. 

\subsection{Analysis of training schemes}
\label{subsec:cold}
A initial subset is needed to start the iterative AL process.
It has two main impacts on the $FT$ scheme as seen in Figure~\ref{fig:acc}. First, at low budgets the fine-tuned model fails to provide strong probability estimates for the acquisition function . This is evident from the results where none of the tested AFs is able to conclusively outperform $random$ sampling. 
It is also the case for $badge$ which has shown improvement over $random$ in ~\cite{ash2020deep}. The key difference is the lower budget setting studied in our work.

Second, the comparative analysis of the two training schemes shows that $FT$ is largely outperformed by the transfer learning strategy for low AL budgets.
The performance of $FT$ scheme starts at 11.16, 13.19 and 32.65 percentage points for balanced $Cifar100$, $Food-10$ and $IMN-100$ respectively. The corresponding accuracy with $SVM$ is significantly higher, with 25.5, 17.54 and 47.29 percentage points respectively. 
This is somewhat intuitive since deep model can be accurately trained only if a relatively large amount of data is available. 
$FT$ lags behind even at the end of the AL process for $CIFAR100$ and $IMN-100$ but becomes competitive for $Food-101$.
This last result is explained by the lower similarity between $Food-101$ and $ILSVRC$ compared to the other datasets.
The efficiency of transfer learning is lower for this dataset, but still much better than fine-tuning in early phase AL.

The comparison of $FT$ and $SVM$ schemes has practical implications for AL.
The training process is much quicker with $SVM$ since it only requires an update of the shallow classifiers. 
Further, a cross-validation step can be envisaged to switch from $SVM$ to $FT$ training when $FT$ outperforms $SVM$ training scheme. 
As suggested by $Food-101$ results, this happens once there are enough samples for a competitive training of deep models.

\subsection{Impact on imbalanced datasets}
The performance of the methods on imbalanced dataset for $SVM$ scheme of annotated subset is presented in Figure~\ref{fig:acc_im}. 
Both $alamp$ and $alamp-div$ provide improvement over the baselines methods. $alamp$ provides average gain of 1.99, 0.9 and 1.49 points for $Cifar100$, $Food-101$ and $MIT-67$ respectively. 
The diversification component is particularly more effective for imbalanced datasets with gains of 3.59, 1.14 and 2.71 points respectively. For example, 50 percent performance on $Cifar100$, is reached with 1600 samples for $alamp-div$, while it takes atleast 2000 samples for any other best method. 
A possible explanation can be found in the imbalance profile of selected subsets (Figure~\ref{fig:ip_im}). The imbalance profiles show the effectiveness of the methods $alamp$ and $alamp-div$ to mitigate the imbalance from being propagated to labeled subset. The results are reported after the use of effective techniques from imbalanced learning. The improvements with the proposed methods also shows the importance of tackling imbalance at the time of sample selection for imbalanced datasets. 

\subsection{Impact of diversification}

The diversification procedure is effective for both balanced and imbalanced dataset, where $alamp-div$ improves results over $alamp$. 
The key reason for its effectiveness is that $alamp$ prioritizes samples with high certainty in the previous iteration. 
Even though the class prediction changes after the update of the model, samples having  different class prediction with high confidence at the previous iteration are likely belong to different regions of representation space. 

We also test the diversification procedure for standard acquisition functions $rand$ and $marg$ on $Cifar100$. $core$ is not considered here as it already selects representative samples and also is not competitive with other AFs. The pseudo class for $rand$ and $marg$ is assigned using the current class prediction. The diversification results are presented in Figure~\ref{fig:div} , with $rand-div$ and $marg-div$ with diversified version of $rand$ and $marg$ respectively. 

$alamp-div$ still provides the best performance, but interestingly $rand-div$ outperforms $rand$. The gain for $rand-div$ is particularly higher at the start of the iterative cycle, where representative sampling is shown to be more important.  
$marg-div$ has very little effect compared to $marg$. This is expected since $marg$ sorts the samples in terms of uncertainty. Thus, the class predictions are not reliable and the diversification procedure becomes ineffective.

\section{Conclusion}

In our work, a transfer learning approach is explored as an alternative to classical fine tuning approach used in deep AL. We show that $SVM$ classifier over fixed representation prove to be very effective alternative at lower budgets. 

The main contribution of this work is the introduction of two new acquisition functions. 
$alamp$ and $alamp-div$ capture the dynamic nature of probability estimates of iterative AL models. 
They outperform competitive baselines over both balanced and imbalanced image classification datasets.
A diversification component is introduced to combine the objectives of informativeness and representativeness. 
We tested the diversification procedure for random sampling, margin sampling and our proposed method.
The diversified version of $alamp$ is particularly effective compared to the other two sampling methods as $alamp$ provides strong pseudo class predictions using the certainty measure from the previous iteration. 


The result of the proposed informative measure and the diversification procedure is inconclusive for $FT$ training scheme. This could be a result of evolving representation space with fine-tuning of model. In the future, we plan to explore ways to implement the proposed informative measure for fine-tuning scheme using different snapshots during the fine-tuning process. 
Further, larger pre-trained models learned on bigger datasets would allow transfer learning scheme to perform well in more application domains.

\bibliographystyle{unsrt}  
\bibliography{references}

\begin{thebibliography}{10}

\bibitem{Settles10activelearning}
Burr Settles.
\newblock Active learning literature survey.
\newblock Technical report, University of Winsconsin, 2010.

\bibitem{DBLP:conf/cvpr/BeluchGNK18}
William~H. Beluch, Tim Genewein, Andreas N{\"{u}}rnberger, and Jan~M.
  K{\"{o}}hler.
\newblock The power of ensembles for active learning in image classification.
\newblock In {\em 2018 {IEEE} Conference on Computer Vision and Pattern
  Recognition, {CVPR} 2018, Salt Lake City, UT, USA, June 18-22, 2018}, pages
  9368--9377, 2018.

\bibitem{DBLP:conf/aaai/CulottaM05}
Aron Culotta and Andrew McCallum.
\newblock Reducing labeling effort for structured prediction tasks.
\newblock In {\em Proceedings, The Twentieth National Conference on Artificial
  Intelligence and the Seventeenth Innovative Applications of Artificial
  Intelligence Conference, July 9-13, 2005, Pittsburgh, Pennsylvania, {USA}},
  pages 746--751, 2005.

\bibitem{DBLP:conf/icdm/SchefferDW01}
Tobias Scheffer, Christian Decomain, and Stefan Wrobel.
\newblock Mining the web with active hidden markov models.
\newblock In {\em Proceedings of the 2001 {IEEE} International Conference on
  Data Mining, 29 November - 2 December 2001, San Jose, California, {USA}},
  pages 645--646, 2001.

\bibitem{Shannon1948}
Claude~Elwood Shannon.
\newblock A mathematical theory of communication.
\newblock 27(3):379--423, 7 1948.

\bibitem{DBLP:conf/icml/DasguptaH08}
Sanjoy Dasgupta and Daniel~J. Hsu.
\newblock Hierarchical sampling for active learning.
\newblock In {\em Machine Learning, Proceedings of the Twenty-Fifth
  International Conference {(ICML} 2008), Helsinki, Finland, June 5-9, 2008},
  pages 208--215, 2008.

\bibitem{DBLP:conf/icmla/LiGC12}
Xianglin Li, Runqiu Guo, and Jun Cheng.
\newblock Incorporating incremental and active learning for scene
  classification.
\newblock In {\em 11th International Conference on Machine Learning and
  Applications, ICMLA, Boca Raton, FL, USA, December 12-15, 2012. Volume 1},
  pages 256--261, 2012.

\bibitem{DBLP:conf/iclr/SenerS18}
Ozan Sener and Silvio Savarese.
\newblock Active learning for convolutional neural networks: {A} core-set
  approach.
\newblock In {\em 6th International Conference on Learning Representations,
  {ICLR} 2018, Vancouver, BC, Canada, April 30 - May 3, 2018, Conference Track
  Proceedings}, 2018.

\bibitem{ash2020deep}
Jordan~T Ash, Chicheng Zhang, Akshay Krishnamurthy, John Langford, and Alekh
  Agarwal.
\newblock Deep batch active learning by diverse, uncertain gradient lower
  bounds.
\newblock In {\em ICLR}, 2020.

\bibitem{DBLP:journals/pami/ChakrabortyBSPY15}
Shayok Chakraborty, Vineeth~Nallure Balasubramanian, Qian Sun, Sethuraman
  Panchanathan, and Jieping Ye.
\newblock Active batch selection via convex relaxations with guaranteed
  solution bounds.
\newblock {\em {IEEE} Trans. Pattern Anal. Mach. Intell.}, 37(10):1945--1958,
  2015.

\bibitem{DBLP:conf/icml/GalIG17}
Yarin Gal, Riashat Islam, and Zoubin Ghahramani.
\newblock Deep bayesian active learning with image data.
\newblock In {\em Proceedings of the 34th International Conference on Machine
  Learning, {ICML} 2017, Sydney, NSW, Australia, 6-11 August 2017}, pages
  1183--1192, 2017.

\bibitem{gao2020consistency}
Mingfei Gao, Zizhao Zhang, Guo Yu, Sercan~{\"O} Ar{\i}k, Larry~S Davis, and
  Tomas Pfister.
\newblock Consistency-based semi-supervised active learning: Towards minimizing
  labeling cost.
\newblock In {\em European Conference on Computer Vision}, pages 510--526.
  Springer, 2020.

\bibitem{konyushkova2017learning}
Ksenia Konyushkova, Raphael Sznitman, and Pascal Fua.
\newblock Learning active learning from data.
\newblock In {\em Advances in Neural Information Processing Systems}, pages
  4225--4235, 2017.

\bibitem{DBLP:conf/cvpr/ZhouSZGGL17}
Zongwei Zhou, Jae~Y. Shin, Lei Zhang, Suryakanth~R. Gurudu, Michael~B. Gotway,
  and Jianming Liang.
\newblock Fine-tuning convolutional neural networks for biomedical image
  analysis: Actively and incrementally.
\newblock In {\em 2017 {IEEE} Conference on Computer Vision and Pattern
  Recognition, {CVPR} 2017, Honolulu, HI, USA, July 21-26, 2017}, pages
  4761--4772, 2017.

\bibitem{Aggarwal_2020_WACV}
Umang Aggarwal, Adrian Popescu, and Celine Hudelot.
\newblock Active learning for imbalanced datasets.
\newblock In {\em Proceedings of the IEEE/CVF Winter Conference on Applications
  of Computer Vision (WACV)}, March 2020.

\bibitem{tong2001support}
Simon Tong and Edward Chang.
\newblock Support vector machine active learning for image retrieval.
\newblock In {\em Proceedings of the ninth ACM international conference on
  Multimedia}, pages 107--118, 2001.

\bibitem{brinker2003incorporating}
Klaus Brinker.
\newblock Incorporating diversity in active learning with support vector
  machines.
\newblock In {\em Proceedings of the 20th international conference on machine
  learning (ICML-03)}, pages 59--66, 2003.

\bibitem{attenberg2013class}
Josh Attenberg and Seyda Ertekin.
\newblock Class imbalance and active learning.
\newblock {\em Imbalanced Learning: Foundations, Algorithms, and Applications},
  pages 101--149, 2013.

\bibitem{ertekin2007active}
Seyda Ertekin, Jian Huang, and C~Lee Giles.
\newblock Active learning for class imbalance problem.
\newblock In {\em Proceedings of the 30th annual international ACM SIGIR
  conference on Research and development in information retrieval}, pages
  823--824, 2007.

\bibitem{zhu2007active}
Jingbo Zhu and Eduard Hovy.
\newblock Active learning for word sense disambiguation with methods for
  addressing the class imbalance problem.
\newblock In {\em Proceedings of the 2007 Joint Conference on Empirical Methods
  in Natural Language Processing and Computational Natural Language Learning
  (EMNLP-CoNLL)}, pages 783--790, 2007.

\bibitem{han2016local}
Lei Han, Kean~Ming Tan, Ting Yang, and Tong Zhang.
\newblock Local uncertainty sampling for large-scale multi-class logistic
  regression.
\newblock {\em arXiv preprint arXiv:1604.08098}, 2016.

\bibitem{hsu2015active}
Wei-Ning Hsu and Hsuan-Tien Lin.
\newblock Active learning by learning.
\newblock In {\em Twenty-Ninth AAAI conference on artificial intelligence}.
  Citeseer, 2015.

\bibitem{wei2015submodularity}
Kai Wei, Rishabh Iyer, and Jeff Bilmes.
\newblock Submodularity in data subset selection and active learning.
\newblock In {\em International Conference on Machine Learning}, pages
  1954--1963, 2015.

\bibitem{Gudovskiy_2020_CVPR}
Denis Gudovskiy, Alec Hodgkinson, Takuya Yamaguchi, and Sotaro Tsukizawa.
\newblock Deep active learning for biased datasets via fisher kernel
  self-supervision.
\newblock In {\em Proceedings of the IEEE/CVF Conference on Computer Vision and
  Pattern Recognition (CVPR)}, June 2020.

\bibitem{berthelot2019mixmatch}
David Berthelot, Nicholas Carlini, Ian Goodfellow, Nicolas Papernot, Avital
  Oliver, and Colin~A Raffel.
\newblock Mixmatch: A holistic approach to semi-supervised learning.
\newblock In {\em Advances in Neural Information Processing Systems}, pages
  5049--5059, 2019.

\bibitem{yunweight}
Juseung Yun, Byungjoo Kim, and Junmo Kim.
\newblock Weight decay scheduling and knowledge distillation for active
  learning.

\bibitem{Coleman2020Selection}
Cody Coleman, Christopher Yeh, Stephen Mussmann, Baharan Mirzasoleiman, Peter
  Bailis, Percy Liang, Jure Leskovec, and Matei Zaharia.
\newblock Selection via proxy: Efficient data selection for deep learning.
\newblock In {\em International Conference on Learning Representations}, 2020.

\bibitem{DBLP:conf/cvpr/HeZRS16}
Kaiming He, Xiangyu Zhang, Shaoqing Ren, and Jian Sun.
\newblock Deep residual learning for image recognition.
\newblock In {\em Conference on Computer Vision and Pattern Recognition}, CVPR,
  2016.

\bibitem{DBLP:journals/nn/BudaMM18}
Mateusz Buda, Atsuto Maki, and Maciej~A. Mazurowski.
\newblock A systematic study of the class imbalance problem in convolutional
  neural networks.
\newblock {\em Neural Networks}, 106:249--259, 2018.

\bibitem{DBLP:journals/ijcv/RussakovskyDSKS15}
Olga Russakovsky, Jia Deng, Hao Su, Jonathan Krause, Sanjeev Satheesh, Sean Ma,
  Zhiheng Huang, Andrej Karpathy, Aditya Khosla, Michael~S. Bernstein,
  Alexander~C. Berg, and Fei{-}Fei Li.
\newblock Imagenet large scale visual recognition challenge.
\newblock {\em International Journal of Computer Vision}, 115(3):211--252,
  2015.

\bibitem{Krizhevsky09learningmultiple}
Alex Krizhevsky.
\newblock Learning multiple layers of features from tiny images.
\newblock Technical report, 2009.

\bibitem{bossard14}
Lukas Bossard, Matthieu Guillaumin, and Luc Van~Gool.
\newblock Food-101 -- mining discriminative components with random forests.
\newblock In {\em European Conference on Computer Vision}, 2014.

\bibitem{DBLP:conf/cvpr/QuattoniT09}
Ariadna Quattoni and Antonio Torralba.
\newblock Recognizing indoor scenes.
\newblock In {\em 2009 {IEEE} Computer Society Conference on Computer Vision
  and Pattern Recognition {(CVPR} 2009), 20-25 June 2009, Miami, Florida,
  {USA}}, pages 413--420, 2009.

\end{thebibliography}

\end{document}